\documentclass{article}

\usepackage[algo2e]{algorithm2e}
\usepackage{microtype}
\usepackage{graphicx}
\usepackage{booktabs} \usepackage{hyperref}

\usepackage{arxiv}

\usepackage{color}
\newcounter{nbdrafts}
\makeatletter
\newcommand{\checknbdrafts}{
\ifnum \thenbdrafts > 0
\@latex@warning@no@line{*WARNING* The document contains \thenbdrafts \space draft note(s)}
\fi}

\makeatother

\usepackage{amsmath}
\usepackage{amssymb}
\usepackage{subcaption}

\newcommand{\linear}{\emph{linear transformer}}
\newcommand{\linears}{\emph{linear transformers}}
\newcommand{\bigO}[1]{\mathcal{O}\left(#1\right)}
\newcommand{\R}[1]{\mathbb{R}^{#1}}
\newcommand{\e}[1]{\exp\left(#1\right)}
\newcommand{\similarity}[1]{\text{sim}\left(#1\right)}
\newcommand{\fe}[1]{\phi\left(#1\right)}
\newcommand{\softmax}[1]{\text{softmax}\left(#1\right)}
\newcommand{\calL}{\mathcal{L}}
\newcommand{\der}[2]{\frac{\partial #1}{\partial #2}}

\icmltitlerunning{Transformers are RNNs}

\begin{document}

\twocolumn[
    \icmltitle{Transformers are RNNs: \\
               Fast Autoregressive Transformers with Linear Attention}

    \icmlsetsymbol{workatidiap}{*}

    \begin{icmlauthorlist}
        \icmlauthor{Angelos Katharopoulos}{idiap,epfl}
        \icmlauthor{Apoorv Vyas}{idiap,epfl}
        \icmlauthor{Nikolaos Pappas}{washington}
        \icmlauthor{Fran\c{c}ois Fleuret}{epfl,unige,workatidiap}
    \end{icmlauthorlist}

    \icmlaffiliation{idiap}{Idiap Research Institute, Switzerland}
    \icmlaffiliation{epfl}{EPFL, Switzerland}
    \icmlaffiliation{washington}{University of Washington, Seattle, USA}
    \icmlaffiliation{unige}{University of Geneva, Switzerland.
                            \textsuperscript{*}Work done at Idiap}

    \icmlcorrespondingauthor{Angelos Katharopoulos}{firstname.lastname@idiap.ch}

    \icmlkeywords{}

    \vskip 0.3in
]

\printAffiliationsAndNotice{}

\begin{abstract}
Transformers achieve remarkable performance in several tasks but due to their
quadratic complexity, with respect to the input's length, they are
prohibitively slow for very long sequences. To address this limitation, we
express the self-attention as a linear dot-product of kernel feature maps and
make use of the associativity property of matrix products to reduce the
complexity from $\bigO{N^2}$ to $\bigO{N}$, where $N$ is the sequence length.
We show that this formulation permits an iterative implementation that
dramatically accelerates autoregressive transformers and reveals their
relationship to recurrent neural networks. Our \textit{linear transformers}
achieve similar performance to vanilla transformers and they are up to 4000x
faster on autoregressive prediction of very long sequences.
\end{abstract}

\section{Introduction}

Transformer models were originally introduced by \citet{vaswani_attn} in
the context of neural machine translation \cite{Sutskever14,bahdanau14} and have
demonstrated impressive results on a variety of tasks dealing with natural
language \cite{devlin-etal-2019}, audio \cite{sperber18interspeech}, and images
\cite{parmar-etal-19}. Apart from tasks with ample supervision, transformers
are also effective in transferring knowledge to tasks with limited or no
supervision when they are pretrained with autoregressive
\cite{Radford2018ImprovingLU,radford2019language} or masked language modeling
objectives \cite{devlin-etal-2019,zhilin19,pmlr-v97-song19d,liu2020roberta}.

However, these benefits often come with a very high computational and memory
cost. The bottleneck is mainly caused by the global receptive field of self-attention, which processes
contexts of $N$ inputs with a quadratic memory and time complexity
$\bigO{N^2}$. As a result, in practice transformers are slow to train and their
context is \textit{limited}. This disrupts temporal coherence and hinders the
capturing of long-term dependencies. \citet{dai-etal-2019-transformer}
addressed the latter by attending to memories from previous contexts albeit at
the expense of computational efficiency.

Lately, researchers shifted their attention to approaches that increase the
context length without sacrificing efficiency. Towards this end,
\citet{child2019generating} introduced sparse factorizations of the attention
matrix to reduce the self-attention complexity to $\bigO{N\sqrt{N}}$.
\citet{kitaev2020reformer} further reduced the complexity to $\bigO{N \log N}$
using locality-sensitive hashing. This made scaling to long sequences
possible. Even though the aforementioned models can be efficiently trained on
large sequences, they do not speed-up autoregressive inference.

In this paper, we introduce the \linear{} model that
significantly reduces the memory footprint and scales linearly with respect to
the context length. We achieve this by using a kernel-based formulation of
self-attention and the associative property of matrix products to calculate the
self-attention weights (\S~\ref{sec:method-linear}). Using our linear
formulation, we also express causal masking with linear complexity and constant
memory (\S~\ref{sec:method-masking}). This reveals the relation between
transformers and RNNs, which enables us to perform autoregressive inference
orders of magnitude faster (\S~\ref{sec:method-trnn}).

Our evaluation on image generation and automatic speech recognition
demonstrates that \linear{} can reach the performance levels
of transformer, while being up to three orders of magnitude faster during
inference.

\section{Related Work}

In this section, we provide an overview of the most relevant works that seek to
address the large memory and computational requirements of transformers.
Furthermore, we discuss methods that theoretically analyze the core
component of the transformer model, namely self-attention. Finally, we present
another line of work that seeks to alleviate the softmax bottleneck in the
attention computation.

\subsection{Efficient Transformers}

Existing works seek to improve memory efficiency in transformers through weight
pruning \cite{paul19}, weight factorization \cite{Lan2020}, weight quantization
\cite{zafrir19} or knowledge distillation. \citet{clark2020electra} proposed a
new pretraining objective called replaced token detection that is more sample
efficient and reduces the overall computation. \citet{lample19} used
product-key attention to increase the capacity of any layer with negligible
computational overhead.

Reducing the memory or computational requirements with these methods leads to
training or inference time speedups, but, fundamentally, the time complexity is
still quadratic with respect to the sequence length which hinders scaling to
long sequences. In contrast, we show that our method reduces both memory
and time complexity of transformers both theoretically
(\S~\ref{sec:method-linear}) and empirically (\S~\ref{sec:synthetic}).

Another line of research aims at increasing the ``context'' of self-attention
in transformers. Context refers to the maximum part of the sequence that is
used for computing self-attention. \citet{dai-etal-2019-transformer} introduced
Transformer-XL which achieves state-of-the-art in language modeling by learning
dependencies beyond a fixed length context without disrupting the temporal
coherence. However, maintaining previous contexts in memory introduces
significant additional computational cost. In contrast,
\citet{sukhbaatar-etal-2019} extended the context length significantly by
learning the optimal attention span per attention head, while maintaining
control over the memory footprint and computation time. Note that both
approaches have the same asymptotic complexity as the vanilla model. In
contrast, we improve the asymptotic complexity of the self-attention, which
allows us to use significantly larger context.

More related to our model are the works of \citet{child2019generating} and
\citet{kitaev2020reformer}. The former \cite{child2019generating} introduced
sparse factorizations of the attention matrix reducing the overall complexity
from quadratic to $\bigO{N\sqrt{N}}$ for generative modeling of long sequences.
More recently, \citet{kitaev2020reformer} proposed Reformer. This method
further reduces complexity to $\bigO{N \log{N}}$ by using locality-sensitive
hashing (LSH) to perform fewer dot products. Note that in order to be able to
use LSH, Reformer constrains the keys, for the attention, to be identical to
the queries. As a result this method cannot be used for decoding tasks where
the keys need to be different from the queries. In comparison, \linears{}
impose no constraints on the queries and keys and scale linearly
with respect to the sequence length. Furthermore, they can be used to perform
inference in autoregressive tasks three orders of magnitude faster, achieving
comparable performance in terms of validation perplexity.

\subsection{Understanding Self-Attention}

There have been few efforts to better understand self-attention from a
theoretical perspective. \citet{tsai2019transformer} proposed a kernel-based
formulation of attention in transformers which considers attention as applying
a kernel smoother over the inputs with the kernel scores being the similarity
between inputs. This formulation provides a better way to understand attention
components and integrate the positional embedding. In contrast, we use the
kernel formulation to speed up the calculation of self-attention and lower its
computational complexity. Also, we observe that if a kernel with positive
similarity scores is applied on the queries and keys, linear attention
converges normally.

More recently, \citet{Cordonnier2020On} provided theoretical proofs and
empirical evidence that a multi-head self-attention with sufficient number of
heads can express any convolutional layer. Here, we instead show that a
self-attention layer trained with an autoregressive objective can be seen as a
recurrent neural network and this observation can be used to significantly
speed up inference time of autoregressive transformer models.

\subsection{Linearized softmax}

For many years, softmax has been the bottleneck for training classification
models with a large number of categories \cite{goodman2001classes,
morin2005hierarchical, mnih2009scalable}. Recent works \cite{blanc2017adaptive,
rawat2019sampled}, have approximated softmax with a linear dot product of
feature maps to speed up the training through sampling. Inspired from these
works, we linearize the softmax attention in transformers. Concurrently with
this work, \citet{shen2020efficient} explored the use of linearized attention for
the task of object detection in images. In comparison, we do not only linearize the
attention computation, but also develop an autoregressive transformer model with
linear complexity and constant memory for both inference and training.
Moreover, we show that through the lens of kernels, every transformer can be
seen as a recurrent neural network.

\section{Linear Transformers}

In this section, we formalize our proposed \linear{}. We present
that changing the attention from the traditional \emph{softmax} attention to a
feature map based dot product attention results in better time and memory
complexity as well as a causal model that can perform sequence generation in
linear time, similar to a recurrent neural network.

Initially, in \S~\ref{sec:method-intro}, we introduce a formulation for the
transformer architecture introduced in \cite{vaswani_attn}. Subsequently, in
\S~\ref{sec:method-linear} and \S~\ref{sec:method-masking} we present our
proposed \linear{} and finally, in \S~\ref{sec:method-trnn} we rewrite the
transformer as a recurrent neural network.

\subsection{Transformers} \label{sec:method-intro}

Let $x \in \R{N \times F}$ denote a sequence of $N$ feature vectors of dimensions
$F$. A transformer is a function $T : \R{N \times F} \to \R{N \times F}$
defined by the composition of $L$ transformer layers $T_1(\cdot), \dots,
T_L(\cdot)$ as follows,
\begin{equation}
    T_l(x) = f_l(A_l(x) + x).
    \label{eq:transformer}
\end{equation}
The function $f_l(\cdot)$ transforms each feature independently of the others
and is usually implemented with a small two-layer feedforward network. $A_l(\cdot)$ is the self
attention function and is the only part of the transformer that acts across
sequences.

The self attention function $A_l(\cdot)$ computes, for every position, a
weighted average of the feature representations of all other positions with a
weight proportional to a similarity score between the representations.
Formally, the input sequence $x$ is projected by three matrices $W_Q \in \R{F
\times D}$, $W_K \in \R{F \times D}$ and $W_V \in \R{F \times M}$ to
corresponding representations $Q$, $K$ and $V$. The output for all
positions, $A_l(x) = V'$, is computed as follows,
\begin{equation}
\begin{aligned}
    Q &= x W_Q, \\
    K &= x W_K,  \\
    V &= x W_V, \\
    A_l(x) = V' &= \softmax{\frac{Q K^T}{\sqrt{D}}} V.
    \label{eq:attn}
\end{aligned}
\end{equation}
Note that in the previous equation, the softmax function is applied rowwise to
$Q K^T$. Following common terminology, the $Q$, $K$ and $V$ are referred to
as the ``queries'', ``keys'' and ``values'' respectively.

Equation \ref{eq:attn} implements a specific form of self-attention called
softmax attention where the similarity score is the exponential of the dot
product between a query and a key. Given that subscripting a matrix with $i$
returns the $i$-th row as a vector, we can write a generalized attention
equation for any similarity function as follows,
\begin{equation}
    V'_i = \frac{\sum_{j=1}^N \similarity{Q_i, K_j} V_j}
                {\sum_{j=1}^N \similarity{Q_i, K_j}}.
    \label{eq:general-attn}
\end{equation}
Equation \ref{eq:general-attn} is equivalent to equation \ref{eq:attn} if we
substitute the similarity function with $\similarity{q, k} = \e{\frac{q^T
k}{\sqrt{D}}}$.

\subsection{Linearized Attention} \label{sec:method-linear}

The definition of attention in equation \ref{eq:attn} is generic and can be
used to define several other attention implementations such as polynomial
attention or RBF kernel attention \cite{tsai2019transformer}. Note that the
only constraint we need to impose to $\similarity{\cdot}$, in order for
equation \ref{eq:general-attn} to define an attention function, is to be
non-negative. This includes all kernels $k(x, y) : \R{2 \times F} \to \R{}_+$.

Given such a kernel with a feature representation $\fe{x}$ we can rewrite
equation \ref{eq:attn} as follows,
\begin{align}
    V'_i = \frac{\sum_{j=1}^N \fe{Q_i}^T \fe{K_j} V_j}
                 {\sum_{j=1}^N \fe{Q_i}^T \fe{K_j}},
\end{align}
and then further simplify it by making use of the associative property of
matrix multiplication to
\begin{align}
     V'_i = \frac{\fe{Q_i}^T \sum_{j=1}^N \fe{K_j} V_j^T}
             {\fe{Q_i}^T \sum_{j=1}^N \fe{K_j}}.
    \label{eq:linear_attn}
\end{align}
The above equation is simpler to follow when the numerator is written in
vectorized form as follows,
\begin{equation}
    \left(\fe{Q} \fe{K}^T\right) V = \fe{Q} \left(\fe{K}^T V\right).
\end{equation}
Note that the feature map $\fe{\cdot}$ is applied rowwise to
the matrices $Q$ and $K$.

From equation \ref{eq:attn}, it is evident that the computational cost of
softmax attention scales with $\bigO{N^2}$, where $N$ represents the sequence
length. The same is true for the memory requirements because the full attention
matrix must be stored to compute the gradients with respect to the queries,
keys and values.
In contrast, our proposed \linear{} from equation \ref{eq:linear_attn} has
time and memory complexity $\bigO{N}$ because we can compute $\sum_{j=1}^N
\fe{K_j} V_j^T$ and $\sum_{j=1}^N \fe{K_j}$ once and reuse them for every
query.

\subsubsection{Feature Maps and Computational Cost} \label{sec:fmap}

For softmax attention, the total cost in terms of multiplications and
additions scales as $\bigO{N^2 \max\left(D, M\right)}$, where $D$ is the
dimensionality of the queries and keys and $M$ is the dimensionality of the
values. On the contrary, for linear attention, we first compute the feature
maps of dimensionality $C$. Subsequently, computing the new values requires
$\bigO{N C M}$ additions and multiplications.

The previous analysis does not take into account the choice of kernel and
feature function. Note that the feature function that corresponds to the
exponential kernel is infinite dimensional, which makes the linearization of
exact softmax attention infeasible. On the other hand, the polynomial kernel,
for example, has an exact finite dimensional feature map and has been shown to
work equally well with the exponential or RBF kernel
\cite{tsai2019transformer}. The computational cost for a linearized polynomial
transformer of degree 2 is $\bigO{N D^2 M}$. This makes the computational
complexity favorable when $N > D^2$. Note that this is true in practice since
we want to be able to process sequences with tens of thousands of elements.

For our experiments, that deal with smaller sequences, we employ a feature map
that results in a positive similarity function as defined below,
\begin{equation}
    \fe{x} = \text{elu}(x) + 1, \label{eq:elu}
\end{equation}
where $\text{elu}(\cdot)$ denotes the exponential linear unit
\cite{clevert2015fast} activation function. We prefer $\text{elu}(\cdot)$ over
$\text{relu}(\cdot)$ to avoid setting the gradients to 0 when $x$ is negative.
This feature map results in an attention function that requires $\bigO{N D M}$
multiplications and additions. In our experimental section, we show that the
feature map of equation \ref{eq:elu} performs on par to the full transformer,
while significantly reducing the computational and memory requirements.

\subsection{Causal Masking} \label{sec:method-masking}

The transformer architecture can be used to efficiently train autoregressive models
by masking the attention computation such that the $i$-th position
can only be influenced by a position $j$ if and only if $j \leq i$, namely a
position cannot be influenced by the subsequent positions. Formally, this
causal masking changes equation \ref{eq:general-attn} as follows,
\begin{equation}
    V'_i = \frac{\sum_{j=1}^i \similarity{Q_i, K_j} V_j}
                {\sum_{j=1}^i \similarity{Q_i, K_j}}.
    \label{eq:masked_attn}
\end{equation}

Following the reasoning of \S~\ref{sec:method-linear}, we linearize the
masked attention as described below,
\begin{equation}
    V'_i = \frac{\fe{Q_i}^T \sum_{j=1}^i \fe{K_j} V_j^T}
                 {\fe{Q_i}^T \sum_{j=1}^i \fe{K_j}}.
    \label{eq:masked_linear_attn}
\end{equation}
By introducing $S_i$ and $Z_i$ as follows,
\begin{align}
    S_i &= \sum_{j=1}^i \fe{K_j} V_j^T, \label{eq:attn_state} \\
    Z_i &= \sum_{j=1}^i \fe{K_j},
\end{align}
we can simplify equation \ref{eq:masked_linear_attn} to
\begin{equation}
    V'_i = \frac{\fe{Q_i}^T S_i}
                {\fe{Q_i}^T Z_i}.
    \label{eq:masked_linear_attn2}
\end{equation}
Note that, $S_i$ and $Z_i$ can be computed from $S_{i-1}$ and $Z_{i-1}$ in
constant time hence making the computational complexity of linear transformers
with causal masking linear with respect to the sequence length.

\subsubsection{Gradient Computation}

A naive implementation of equation \ref{eq:masked_linear_attn2}, in any deep
learning framework, requires storing all intermediate values $S_i$ in order to
compute the gradients. This increases the memory consumption by $\max\left(D,
M\right)$ times; thus hindering the applicability of causal linear attention to
longer sequences or deeper models. To address this, we derive the gradients of the numerator in
equation \ref{eq:masked_linear_attn} as cumulative sums. This allows us to
compute both the forward and backward pass of causal linear attention in
\textbf{linear time} and \textbf{constant memory}. A detailed derivation is
provided in the supplementary material.

Given the numerator $\bar{V}_i$ and the gradient of a scalar loss function with
respect to the numerator $\nabla_{\bar{V}_i}\calL$, we derive
$\nabla_{\fe{Q_i}}\calL$, $\nabla_{\fe{K_i}}\calL$ and $\nabla_{V_i}\calL$ as follows,
\begin{align}
    \nabla_{\fe{Q_i}} \calL &= \nabla_{\bar{V}_i}\calL
        \left(\sum_{j=1}^i \fe{K_j} V_j^T\right)^T,
        \label{eq:grad_q} \\
    \nabla_{\fe{K_i}}\calL &= \left(\sum_{j=i}^N \fe{Q_j}
        \left(\nabla_{\bar{V}_j}\calL\right)^T\right) V_i\ ,
        \label{eq:grad_k} \\
    \nabla_{V_i}\calL &= \left(\sum_{j=i}^N \fe{Q_j}
        \left(\nabla_{\bar{V}_j}\calL\right)^T\right)^T \fe{K_i}.
        \label{eq:grad_v}
\end{align}
The cumulative sum terms in equations \ref{eq:masked_linear_attn},
\ref{eq:grad_q}-\ref{eq:grad_v} are computed in linear time and require
constant memory with respect to the sequence length. This results in an
algorithm with computational complexity $\bigO{N C M}$ and memory $\bigO{N
\max\left(C, M\right)}$ for a given feature map of $C$ dimensions. A pseudocode
implementation of the forward and backward pass of the numerator is given in
algorithm \ref{alg:causal_product}.

\subsubsection{Training and Inference}

When training an autoregressive transformer model the full ground truth
sequence is available. This makes layerwise parallelism possible both for
$f_l(\cdot)$ of equation \ref{eq:transformer} and the attention computation.
As a result, transformers are more efficient to train than recurrent neural
networks. On the other hand, during inference the output for timestep $i$ is
the input for timestep $i+1$. This makes autoregressive models impossible to
parallelize. Moreover, the cost per timestep for transformers is not constant;
instead, it scales with the square of the current sequence length because
attention must be computed for all previous timesteps.

Our proposed \linear{} model \emph{combines the best of both worlds}.
When it comes to training, the computations can be parallelized and take full
advantage of GPUs or other accelerators. When it comes to inference, the cost
per time and memory for one prediction is constant for our model. This means we
can simply store the $\fe{K_j}V_j^T$ matrix as an internal state and update it
at every time step like a recurrent neural network. This results in inference
\textbf{thousands of times faster} than other transformer models.

\subsection{Transformers are RNNs} \label{sec:method-trnn}

In literature, transformer models are considered to be a fundamentally
different approach to recurrent neural networks. However, from the causal
masking formulation in \S~\ref{sec:method-masking} and the discussion in the
previous section, it becomes evident that any transformer layer with causal
masking can be written as a model that, given an input, modifies an internal
state and then predicts an output, namely a Recurrent Neural Network (RNN). Note that, in contrast to
Universal Transformers \cite{dehghani2018universal}, we consider the recurrence
with respect to time and not depth.

In the following equations, we formalize the transformer layer of equation
\ref{eq:transformer} as a recurrent neural network. The resulting RNN
has two hidden states, namely the attention memory $s$ and the
normalizer memory $z$. We use subscripts to denote the timestep in the
recurrence.
\begin{align}
    s_0 &= 0 \label{eq:trnn-1}, \\
    z_0 &= 0 \label{eq:trnn-2}, \\
    s_i &= s_{i-1} + \fe{x_i W_K} \left(x_i W_V\right)^T, \label{eq:trnn-3} \\
    z_i &= z_{i-1} + \fe{x_i W_K}, \label{eq:trnn-4} \\
    y_i &= f_l\left(\frac{\fe{x_i W_Q}^T s_i}{\fe{x_i W_Q}^T z_i} + x_i\right).
    \label{eq:trnn-5}
\end{align}
In the above equations, $x_i$ denotes the $i$-th input and $y_i$ the $i$-th
output for a specific transformer layer. Note that our formulation does not
impose any constraint on the feature function and it can be used for
representing \emph{any transformer} model, in theory even the ones using softmax
attention. This formulation is a first step towards better understanding the
relationship between transformers and popular recurrent networks
\cite{hochreiter1997long} and the processes used for storing and retrieving
information.

\section{Experiments}

\begin{algorithm}[h]
    \caption{Linear transformers with causal masking} \label{alg:causal_product}
    \DontPrintSemicolon
    \SetKwFunction{forward}{forward}
    \SetKwFunction{backward}{backward}
    \SetKwProg{Fn}{function}{:}{end}
    \Fn{\forward{$\fe{Q}$, $\fe{K}$, $V$}}{
        $V' \gets 0$,
        $S \gets 0$ \\
        \For{$i = 1, \dots, N$}{
            $S \gets S + \fe{K_i} V_i^T$ \hfill equation \ref{eq:attn_state} \\
            $\bar{V}_i \gets \fe{Q_i} S$
        }
        \KwRet $\bar{V}$
    }
    \;
    \Fn{\backward{$\fe{Q}$, $\fe{K}$, $V$, $G$}}{
        \tcc{$G$ is the gradient of the loss with respect to the output of
             \forward}
        $S \gets 0$,
        $\nabla_{\fe{Q}}\calL \gets 0$ \\
        \For{$i = 1, \dots, N$}{
            $S \gets S + \fe{K_i} V_i^T$ \\
            $\nabla_{\fe{Q_i}}\calL \gets G_i S^T$ \hfill equation \ref{eq:grad_q} \\
        }
        $S \gets 0$,
        $\nabla_{\fe{K}}\calL \gets 0$,
        $\nabla_{V}\calL \gets 0$ \\
        \For{$i = N, \dots, 1$}{
            $S \gets S + \fe{Q_i} G_i^T$ \\
            $\nabla_{V_i}\calL \gets S^T \fe{K_i}$ \hfill equation \ref{eq:grad_v} \\
            $\nabla_{\fe{K_i}}\calL \gets S V_i$ \hfill equation \ref{eq:grad_k}
        }
        \KwRet $\nabla_{\fe{Q}}\calL$, $\nabla_{\fe{K}}\calL$, $\nabla_{V}\calL$
    }
\end{algorithm}

In this section, we analyze experimentally the performance of the proposed
\linear{}. Initially, in \S~\ref{sec:synthetic}, we evaluate the
linearized attention in terms of computational cost, memory consumption and
convergence on synthetic data. To further showcase the effectiveness of
\linears{}, we evaluate our model on two real-world applications,
image generation in \S~\ref{sec:image} and automatic speech recognition in
\S~\ref{sec:asr}. We show that our model achieves competitive performance with respect to the
state-of-the-art transformer architectures, while requiring significantly less
GPU memory and computation.

Throughout our experiments, we compare our model with two baselines, the full
transformer with softmax attention and the Reformer \cite{kitaev2020reformer},
the latter being a state-of-the-art accelerated transformer architecture. For
the Reformer, we use a PyTorch reimplementation of the published code
and for the full transformer we use the default PyTorch implementation.
Note that for Reformer, we do not use the reversible layers, however,
this does not affect the results as we only measure the memory consumption with
respect to the self attention layer. In all experiments, we use
\textbf{softmax} \cite{vaswani_attn} to refer to the standard transformer
architecture, \textbf{linear} for our proposed \linears{} and
\textbf{lsh-X} for Reformer \cite{kitaev2020reformer}, where \emph{X} denotes
the hashing rounds.

For training the \linears{}, we use the feature map of equation
\ref{eq:elu}. Our PyTorch \cite{paszke2019pytorch} code with documentation and
examples can be found at \url{https://linear-transformers.com/}. The constant
memory gradient computation of equations \ref{eq:grad_q}-\ref{eq:grad_v} is
implemented in approximately 200 lines of CUDA code.

\begin{figure*}
    \centering
    \begin{subfigure}[t]{\columnwidth}
        \includegraphics[width=\linewidth]{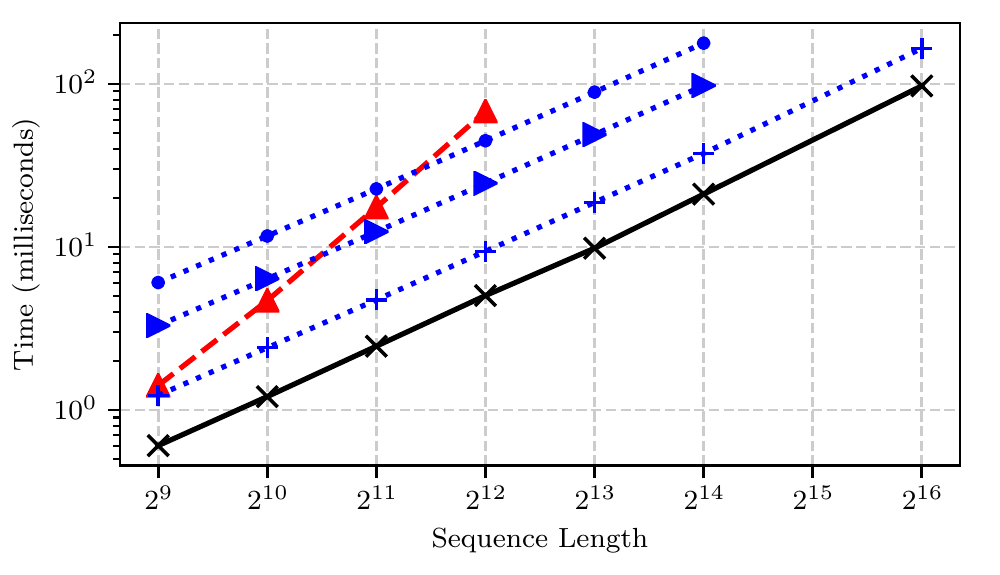}
    \end{subfigure}
    \begin{subfigure}[t]{\columnwidth}
        \includegraphics[width=\linewidth]{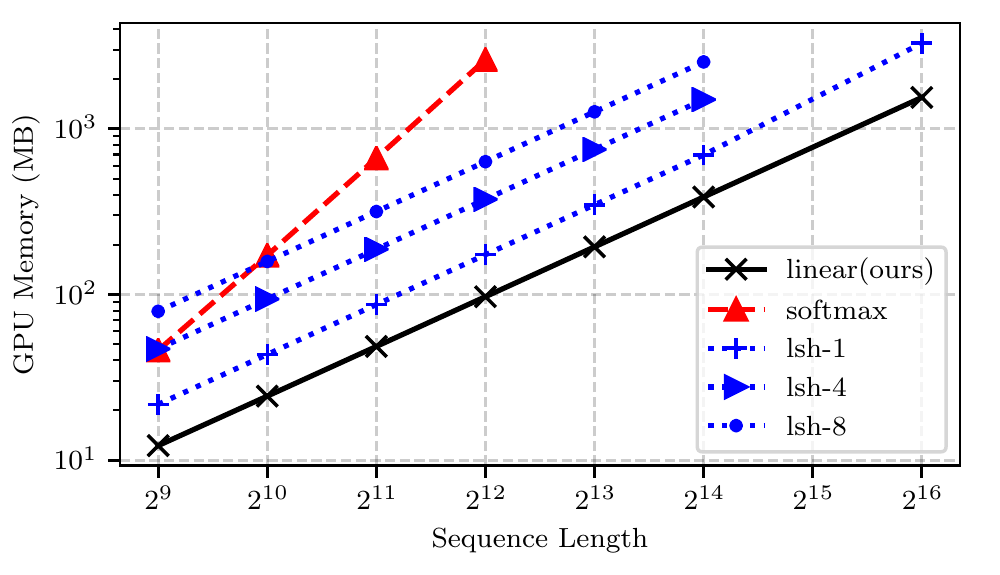}
    \end{subfigure}
    \caption{Comparison of the computational requirements for a forward/backward pass for
             Reformer (lsh-X), softmax attention and linear attention. Linear
             and Reformer models scale linearly with the sequence length unlike
             softmax which scales with the square of the sequence length both
             in memory and time. Full details of the experiment can be found in
             \S~\ref{sec:synthetic}.}
    \label{fig:benchmark}
\end{figure*}

\begin{figure}
    \centering
    \includegraphics[width=\linewidth]{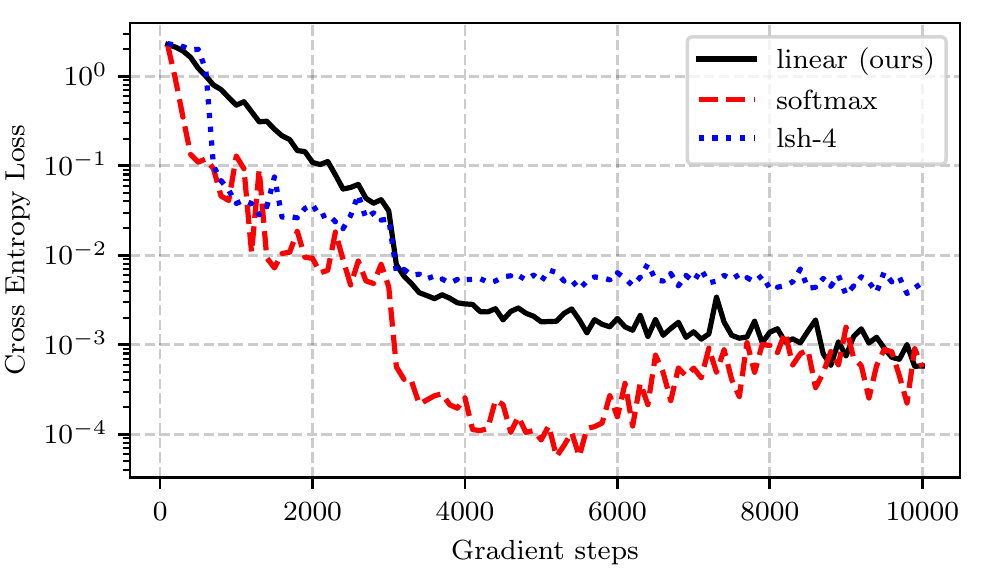}
    \caption{Convergence comparison of \emph{softmax}, \emph{linear} and
             \emph{reformer} attention on a sequence duplication task.
             \emph{linear} converges stably and reaches the same final
             performance as softmax. The details of the experiment are in
             \S~\ref{sec:synthetic}.}
    \label{fig:causal-copy}
\end{figure}

\subsection{Synthetic Tasks} \label{sec:synthetic}

\subsubsection{Convergence Analysis}

To examine the convergence properties of \linears{} we train on an
artifical copy task with causal masking. Namely, the transformers have to copy
a series of symbols similar to the sequence duplication task of \citet{kitaev2020reformer}. We
use a sequence of maximum length 128 with 10 different symbols separated by a
dedicated separator symbol. For all three methods, we train a 4 layer
transformer with 8 attention heads using a batch size of 64 and the RAdam
optimizer \cite{liu2019variance} with a learning rate of $10^{-3}$ which is
reduced to $10^{-4}$ after 3000 updates. Figure \ref{fig:causal-copy} depicts
the loss with respect to the number of gradient steps. We observe that
linear converges smoothly and reaches a lower loss than lsh due to
the lack of noise introduced by hashing. In particular, it reaches
the same loss as softmax.

\subsubsection{Memory and Computational Requirements}

In this subsection, we compare transformers with respect to their computational
and memory requirements. We compute the attention and the gradients for a
synthetic input with varying sequence lengths $N \in \{2^9, 2^{10}, \dots,
2^{16}\}$ and measure the peak allocated GPU memory and required time for each
variation of transformer.
We scale the batch size inversely with the sequence length and
report the time and memory per sample in the batch.

Every method is evaluated up to the maximum sequence length that fits the GPU
memory. For this benchmark we use an NVidia GTX 1080 Ti with 11GB of memory.
This results in a maximum sequence length of 4,096 elements for softmax and
16,384 for lsh-4 and lsh-8. As expected, softmax scales quadratically with
respect to the sequence length.  Our method is faster and requires less memory
than the baselines for every configuration, as seen in figure
\ref{fig:benchmark}.  We observe that both Reformer and linear attention scale
linearly with the sequence length. Note that although the asymptotic complexity
for Reformer is $\bigO{N \log N}$, $\log N$ is small enough and does not affect
the computation time.

\subsection{Image Generation} \label{sec:image}

Transformers
have shown great results on the task of conditional or unconditional
autoregressive generation \cite{radford2019language, child2019generating},
however, sampling from transformers is slow due to the task being inherently
sequential and the memory scaling with the square of the sequence length.
In this section, we train causally masked transformers to predict images
pixel by pixel. Our achieved performance in terms of bits
per dimension is on par with \emph{softmax} attention while being able to generate
images \textbf{more than 1,000 times faster} and with \textbf{constant memory
per image} from the first to the last pixel. We refer the reader to our
supplementary for comparisons in terms of training evolution, quality of
generated images and time to generate a single image. In addition, we also
compare with a faster softmax transformer that caches the keys and values
during inference, in contrast to the PyTorch implementation.

\subsubsection{MNIST} \label{sec:exp-mnist}

\bgroup
\renewcommand{\arraystretch}{1.1}
\begin{table}
    \begin{center}
    \begin{tabular}{lcrl}
        Method & Bits/dim & \multicolumn{2}{c}{Images/sec}\\
        \hline
        Softmax & 0.621 & 0.45 & (1$\times$) \\
        LSH-1 & 0.745 & 0.68 & (1.5$\times$) \\
        LSH-4 & 0.676 & 0.27 & (0.6$\times$) \\
        \hline
        Linear (ours) & 0.644 & \textbf{142.8} & \textbf{(317$\times$)}
    \end{tabular}
    \end{center}
    \caption{Comparison of autoregressive image generation of MNIST images. Our
             linear transformers achieve almost the same bits/dim as the full
             softmax attention but more than 300 times higher throughput in
             image generation. The full details of the experiment are in
             \S~\ref{sec:exp-mnist}.}
    \label{tab:mnist}
\end{table}
\egroup

First, we evaluate our model on image generation with autoregressive
transformers on the widely used MNIST dataset \cite{lecun2010mnist}. The
architecture for this experiment comprises 8 attention layers with 8 attention
heads each. We set the embedding size to 256 which is 32 dimensions per head.
Our feed forward dimensions are 4 times larger than our embedding size. We
model the output with a mixture of 10 logistics as introduced by
\citet{salimans2017pixelcnn++}. We use the RAdam optimizer with a learning rate
of $10^{-4}$ and train all models for 250 epochs. For the reformer baseline, we
use 1 and 4 hashing rounds. Furthermore, as suggested in
\citet{kitaev2020reformer}, we use 64 buckets and chunks with approximately 32
elements. In particular, we divide the 783 long input sequence to 27 chunks of
29 elements each. Since the sequence length is realtively small, namely only
784 pixels, to remove differences due to different batch sizes we use a batch
size of 10 for all methods.

Table \ref{tab:mnist} summarizes the results. We observe that linear
transformers achieve almost the same performance, in terms of final perplexity,
as softmax transformers while being able to generate images more than 300 times
faster. This is achieved due to the low memory requirements of our model, which
is able to simultaneously generate 10,000 MNIST images with a single GPU. In
particular, the memory is constant with respect to the sequence length because
the only thing that needs to be stored between pixels are the $s_i$ and $z_i$
values as described in equations \ref{eq:trnn-3} and \ref{eq:trnn-4}. On the
other hand, both softmax and Reformer require memory that increases with the
length of the sequence.

\begin{figure}[h]
    \centering
    \begin{subfigure}[t]{0.9\columnwidth}
        \centering
        Unconditional samples
    \end{subfigure}
    \begin{subfigure}[t]{0.9\columnwidth}
        \includegraphics[width=\linewidth]{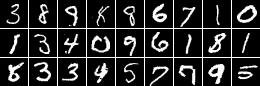}
        \vspace{0.3em}
    \end{subfigure}
    \begin{subfigure}[t]{0.9\columnwidth}
        \centering
        Image completion
    \end{subfigure}
    \begin{subfigure}[t]{0.0971\columnwidth}
        \includegraphics[width=\linewidth]{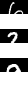}
        \caption{}
    \end{subfigure}\,\,
    \begin{subfigure}[t]{0.6\columnwidth}
        \includegraphics[width=\linewidth]{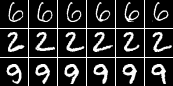}
        \caption{}
    \end{subfigure}\,\,
    \begin{subfigure}[t]{0.0971\columnwidth}
        \includegraphics[width=\linewidth]{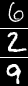}
        \caption{}
    \end{subfigure}
    \caption{Unconditional samples and image completions generated by our
             method for MNIST. (a) depicts the occluded orignal images, (b) the
             completions and (c) the original. Our model achieves comparable
             bits/dimension to softmax, while having more than
             \textbf{300 times} higher throughput, generating \textbf{142
             images/second}. For details see \S~\ref{sec:exp-mnist}.}
    \label{fig:mnist-images}
\end{figure}

Image completions and unconditional samples from our MNIST model can be seen in
figure \ref{fig:mnist-images}. We observe that our linear transformer generates
very convincing samples with sharp boundaries and no noise. In the
case of image completion, we also observe that the transformer learns to use
the same stroke style and width as the original image effectively attending
over long temporal distances. Note that as the achieved perplexity is more or
less the same for all models, we do not observe qualitative differences between
the generated samples from different models.

\subsubsection{CIFAR-10} \label{sec:exp-cifar}

\bgroup
\renewcommand{\arraystretch}{1.1}
\begin{table}
    \begin{center}
    \begin{tabular}{lcrl}
        Method & Bits/dim & \multicolumn{2}{c}{Images/sec} \\
        \hline
        Softmax & 3.47 & 0.004 & (1$\times$) \\
        LSH-1 & 3.39 & 0.015  & (3.75$\times$) \\
        LSH-4 & 3.51 & 0.005  & (1.25$\times$) \\
        \hline
        Linear (ours) & 3.40 & \textbf{17.85} & \textbf{(4,462$\times$)}
    \end{tabular}
    \end{center}
    \caption{We train autoregressive transformers for 1 week on a single GPU to
             generate CIFAR-10 images. Our linear transformer completes 3 times
             more epochs than softmax, which results in better perplexity. Our model
             generates images 4,000$\times$ faster than the baselines. The
             full details of the experiment are in \S~\ref{sec:exp-cifar}.}
    \label{tab:cifar10}
\end{table}
\egroup

The benefits of our linear formulation increase as the sequence length increases.
To showcase that, we train 16 layer transformers to generate
CIFAR-10 images \cite{krizhevsky2009learning}. For each layer we use the same
configuration as in the previous experiment. For Reformer, we use again 64
buckets and 83 chunks of 37 elements, which is approximately 32, as suggested in
the paper. Since the sequence length is almost 4 times larger than for the
previous experiment, the full transformer can only be used with a batch size of 1
in the largest GPU that is available to us, namely an NVidia P40 with 24GB of
memory. For both the linear transformer and reformer, we use a batch size
of 4. All models are trained for 7 days. We report results in terms of
bits per dimension and image generation throughput in table \ref{tab:cifar10}.
Note that although the main point of this experiment is not the final
perplexity, it is evident that as the sequence length grows, the fast
transformer models become increasingly more efficient per GPU hour, achieving
better scores than their slower counterparts.

\begin{figure}[h]
    \centering
    \begin{subfigure}[t]{0.9\columnwidth}
        \centering
        Unconditional samples
    \end{subfigure}
    \begin{subfigure}[t]{0.9\columnwidth}
        \includegraphics[width=\linewidth]{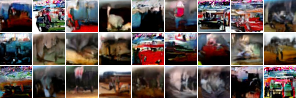}
        \vspace{0.3em}
    \end{subfigure}
    \begin{subfigure}[t]{0.9\columnwidth}
        \centering
        Image completion
    \end{subfigure}
    \begin{subfigure}[t]{0.0971\columnwidth}
        \includegraphics[width=\linewidth]{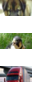}
        \caption{}
    \end{subfigure}\,\,
    \begin{subfigure}[t]{0.6\columnwidth}
        \includegraphics[width=\linewidth]{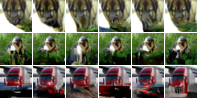}
        \caption{} \label{fig:cifar-completions}
    \end{subfigure}\,\,
    \begin{subfigure}[t]{0.0971\columnwidth}
        \includegraphics[width=\linewidth]{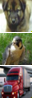}
        \caption{}
    \end{subfigure}
    \caption{Unconditional samples and image completions generated by our
             method for CIFAR-10. (a) depicts the occluded orignal images, (b) the
             completions and (c) the original. As the sequence length grows
             linear transformers become more efficient compared to softmax
             attention. Our model achieves more than \textbf{4,000 times}
             higher throughput and generates \textbf{17.85 images/second}.
             For details see \S~\ref{sec:exp-cifar}.}
    \label{fig:cifar-images}
\end{figure}

As the memory and time to generate a single pixel scales quadratically with the
number of pixels for both Reformer and softmax attention, the increase in
throughput for our linear transformer is even more pronounced. In particular,
\textbf{for every image generated} by the softmax transformer, \textbf{our
method can generate 4,460 images}. Image completions and unconditional samples
from our model can be seen in figure \ref{fig:cifar-images}. We observe that
our model generates images with spatial consistency and can complete images
convincigly without significantly hindering the recognition of the image
category. For instance, in figure \ref{fig:cifar-completions}, all images have
successfully completed the dog's nose (first row) or the windshield of the
truck (last row).

\subsection{Automatic Speech Recognition} \label{sec:asr}

\bgroup
\renewcommand{\arraystretch}{1.1}
\begin{table}
    \begin{center}
    \begin{tabular}{lrr}
        Method & Validation PER & Time/epoch (s) \\
        \hline
        Bi-LSTM & 10.94 & 1047 \\
        \hline
        Softmax &  5.12 & 2711 \\
               LSH-4   &  9.33 & 2250 \\
        \hline
        Linear (ours) & 8.08 & \textbf{824}
    \end{tabular}
    \end{center}
    \caption{Performance comparison in automatic speech recognition on the WSJ
             dataset. The results are given in the form of phoneme error rate
             (PER) and training time per epoch. Our model outperforms the LSTM
             and Reformer while being faster to train and evaluate. Details of
             the experiment can be found in \S~\ref{sec:asr}.}
    \label{tab:asr}
    \vspace{-1em}
\end{table}
\egroup

To show that our method can also be used for non-autoregressive tasks, we
evaluate the performance of linear transformers in end-to-end automatic speech
recognition using Connectionist Temporal Classification (CTC) loss
\cite{graves2006connectionist}. In this setup, we predict a distribution over
phonemes for each input frame in a non autoregressive fashion. We use the 80
hour WSJ dataset \cite{paul1992design} with 40-dimensional mel-scale
filterbanks without temporal differences as features. The dataset contains
sequences with 800 frames on average and a maximum sequence length of 2,400
frames.
For this task, we also compare with a bidirectional LSTM
\cite{hochreiter1997long} with 3 layers of hidden size 320. We use the Adam
optimizer \cite{kingma2014adam} with a learning rate of $10^{-3}$ which is
reduced when the validation error stops decreasing. For the transformer models,
we use 9 layers with 6 heads with the same embedding dimensions as for the
image experiments. As an optimizer, we use RAdam with an initial learning rate
of $10^{-4}$ that is divided by 2 when the validation error stops decreasing.

All models are evaluated in terms of phoneme error rate (PER) and training time per
epoch. We observe that linear outperforms the recurrent network baseline and
Reformer both in terms of performance and speed by a large margin, as seen in
table \ref{tab:asr}. Note that the softmax transformer, achieves lower phone
error rate in comparison to all baselines, but is significantly slower. In
particular, \linear{} is more than $3 \times$ faster per epoch. We provide
training evolution plots in the supplementary.

\section{Conclusions}

In this work, we presented \linear{}, a model that significantly
reduces the memory and computational cost of the original transformers. In
particular, by exploiting the associativity property of matrix products we are
able to compute the self-attention in time and memory that scales linearly with
respect to the sequence length. We show that our model can be used with causal
masking and still retain its linear asymptotic complexities. Finally, we
express the transformer model as a recurrent neural network, which allows us to
perform inference on autoregressive tasks thousands of time faster.

This property opens a multitude of directions for future research regarding the
storage and retrieval of information in both RNNs and transformers. Another
line of research to be explored is related to the choice of feature map for
linear attention. For instance, approximating the RBF kernel with random
Fourier features could allow us to use models pretrained with softmax
attention.

\section*{Acknowledgements}

Angelos Katharopoulos was supported by the Swiss National Science Foundation
under grant numbers FNS-30209 "ISUL" and FNS-30224 "CORTI". Apoorv Vyas was
supported by the Swiss National Science Foundation under grant number FNS-30213
"SHISSM". Nikolaos Pappas was supported by the Swiss National Science
Foundation under grant number P400P2\_183911 "UNISON".

\bibliographystyle{icml2020}
\bibliography{references}

\appendix
\onecolumn
\standalonetitle{Supplementary Material for \\
                 Transformers are RNNs: \\
                 Fast Autoregressive Transformers with Linear Attention}
\section{Gradient Derivation}

In the first section of our supplementary material, we derive in detail the
gradients for causally masked linear transformers and show that they can be
computed in linear time and constant memory. In particular, we derive the
gradients of a scalar loss with respect to the numerator of the following
equation,
\begin{equation}
    V'_i = \frac{\fe{Q_i}^T \sum_{j=1}^i \fe{K_j} V_j^T}
                {\fe{Q_i}^T \sum_{j=1}^i \fe{K_j}}.
\end{equation}
The gradient with respect to the denominator and the fraction are efficiently
handled by autograd. Without loss of generality, we can assume that $Q$ and $K$
already contain the vectors mapped by $\fe{\cdot}$, hence given the numerator
\begin{equation}
    \bar{V}_i = Q_i^T \sum_{j=1}^i K_j V_j^T,
\end{equation}
and $\nabla_{\bar{V}}\calL$ we seek to compute $\nabla_{Q}\calL$,
$\nabla_{K}\calL$ and $\nabla_{V}\calL$. Note that $Q \in \R{N \times D}$, $K
\in \R{N \times D}$ and $V \in \R{N \times M}$. To derive the gradients, we
first express the above equation for a single element without using vector notation,
\begin{equation}
    \bar{V}_{ie} =
        \sum_{d=1}^D Q_{id} \sum_{j=1}^i K_{jd} V_{je} =
        \sum_{d=1}^D \sum_{j=1}^i Q_{id} K_{jd} V_{je}.
\end{equation}
Subsequently we can start deriving the gradients for $Q$ by taking the partial
derivative for any $Q_{lt}$, as follows
\begin{equation}
    \der{\calL}{Q_{lt}} = \sum_{e=1}^M \der{\calL}{\bar{V}_{le}} \der{\bar{V}_{le}}{Q_{lt}} =
        \sum_{e=1}^M \der{\calL}{\bar{V}_{le}} \left(
            \sum_{j=1}^l K_{jt} V_{je}
        \right). \label{eq:der_q}
\end{equation}
If we write the above equation as a matrix product of gradients it becomes,
\begin{equation}
    \nabla_{Q_i}\calL = \nabla_{\bar{V}_i}\calL \left(
        \sum_{j=1}^i K_j V_j^T
    \right)^T,
\end{equation}
proving equation 13 from the main paper. In equation~\ref{eq:der_q} we made use
of the fact that $Q_{lt}$ only affects $\bar{V}_{l}$ hence we do not need to
sum over $i$ to compute the gradients. However, for $K$ and $V$ this is not the
case. In particular, $K_j$ affects all $\bar{V}_i$ where $i \geq j$. Consequently,
we can write the partial derivative of the loss with respect to $K_{lt}$ as follows,
\begin{equation}
\begin{aligned}
    \der{\calL}{K_{lt}} &=
        \sum_{e=1}^M \sum_{i=l}^N \der{\calL}{\bar{V}_{ie}} \der{\bar{V}_{ie}}{K_{lt}} =
        \sum_{e=1}^M \sum_{i=l}^N \der{\calL}{\bar{V}_{ie}}
        \der{\left(\sum_{d=1}^D \sum_{j=1}^i Q_{id} K_{jd} V_{je}\right)}{K_{lt}} \\
     &= \sum_{e=1}^M \sum_{i=l}^N \der{\calL}{\bar{V}_{ie}} Q_{it} V_{le}.
\end{aligned}
\end{equation}
As for $Q$ we can now write the gradient in vectorized form,
\begin{equation}
    \nabla_{K_i} \calL = \left(
        \sum_{j=i}^N Q_j \left(\nabla_{\bar{V}_j} \calL\right)^T
        \right) V_i,
\end{equation}
proving equation 14 from the paper. Following the same reasoning, we can
compute the partial derivative of the loss with respect to $V_{lt}$ and prove
equation 15. Note that the cumulative sum matrices for the gradient with
respect to $Q$ and $K$ have the same size, however one is computed in the
forward direction (summing from 1 to $N$) similarly to the forward pass and the
other is computed in the backwards direction (summing from $N$ to 1) similar to
backpropagation through time done in RNNs.

\section{Training Evolution}

\begin{figure*}[h]
    \centering
    \begin{subfigure}[t]{0.33\columnwidth}
        \includegraphics[width=\linewidth]{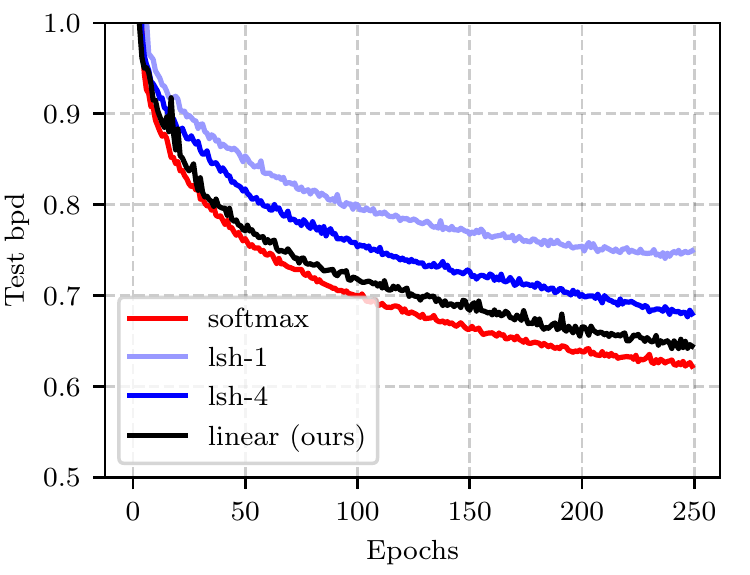}
        \caption{MNIST} \label{fig:evo_mnist}
    \end{subfigure}
    \begin{subfigure}[t]{0.33\columnwidth}
        \includegraphics[width=\linewidth]{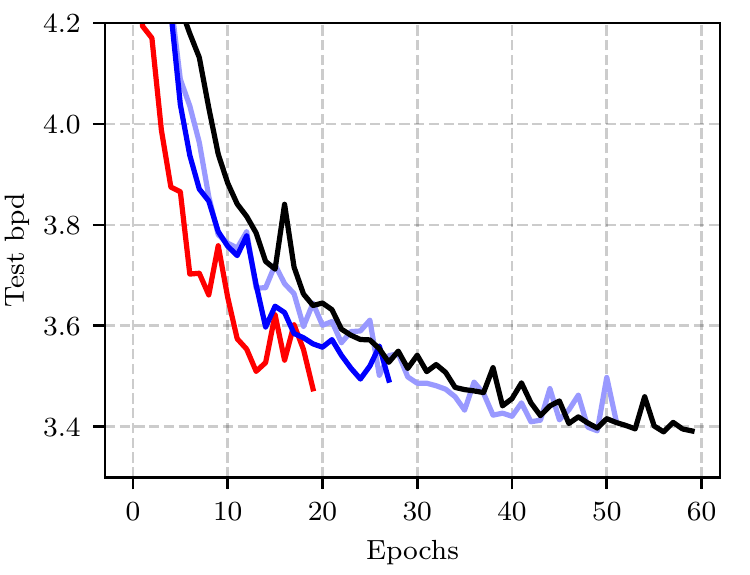}
        \caption{CIFAR-10} \label{fig:evo_cifar}
    \end{subfigure}
    \begin{subfigure}[t]{0.33\columnwidth}
        \includegraphics[width=\linewidth]{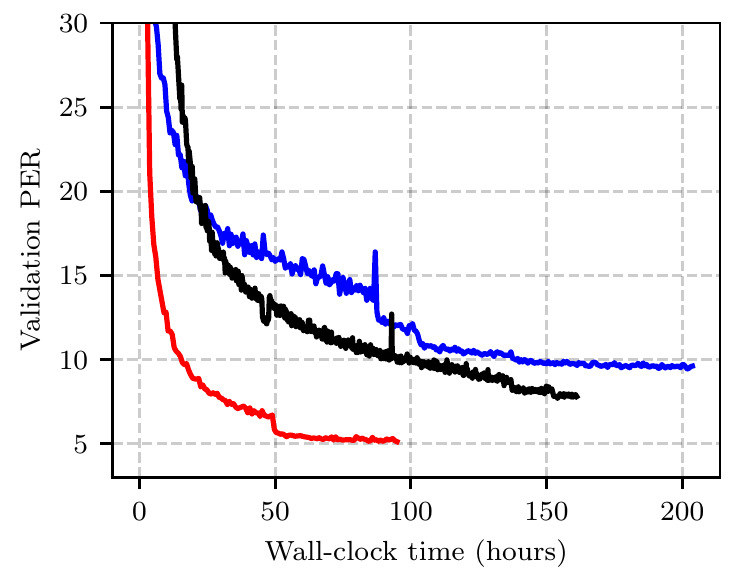}
        \caption{Speech Recognition} \label{fig:evo_asr}
    \end{subfigure}
    \caption{Training evolution of transformers for all our experiments. It can
             be observed that \linears{} converge consistently faster than
             Reformer and in the autoregressive experiments on par with
             softmax. For MNIST all methods are trained for 250 epochs while
             for CIFAR we train for 7 days. In the speech recognition
             experiments all methods are trained to convergence. The details of
             the experiments can be found in \S~4.2.1, \S~4.2.2 and \S~4.3 in
             the main paper.}
    \label{fig:evo}
\end{figure*}

In figure \ref{fig:evo} we present the training evolution of all transformer
models in our experiments. For the MNIST experiment (Fig.~\ref{fig:evo_mnist})
we train all methods for 250 epochs. The sequence length is small enough so
that the training time does not vary significantly for all methods. We observe
that our method converges on par with softmax attention outperforming
significantly both reformer variants.

On the other hand, for CIFAR-10 (Fig.~\ref{fig:evo_cifar}) we train all methods
for a fixed amount of time, namely 7 days. We observe that \emph{lsh-1} and
\emph{linear} complete significantly more epochs than softmax and lsh-4 and
achieve better performance. This gap is expected to increase with a further
increase in sequence length.

Finally, in our last experiment on automatic speech recognition
(Fig.~\ref{fig:evo_asr}), softmax outperforms significantly both Reformer and
linear in terms of convergence. Note that linear is $3\times$ faster per epoch
which means it has completed approximately 4 times more epochs in comparison to
softmax. Even though softmax attention is better in this task, we observe that
\linears{} significantly outperform Reformer both in terms of convergence and
final performance.

\section{Image Generation Throughput Discussion} \label{sec:imgen-discussion}

\subsection{Stateful softmax attention} \label{sec:imgen-discussion-1}

In \S~4.2 of the main paper, we report the image generation
throughput and we compare with \textbf{softmax} transformer and \textbf{lsh}.
In this section we create another baseline, denoted as
\textbf{stateful-softmax}, that implements a softmax autoregressive transformer
as a recurrent model. Namely, all the keys and values are saved and then passed
to the model again when predicting the next element of the sequence. The state
of this recurrent model is the set of keys and values which has size
proportional to the sequence length. This is qualitatively different to our
proposed model that has a state with fixed dimensions and computing the $i$-th
state given the previous one has fixed computational cost regardless of $i$.

\begin{table}
    \begin{subtable}[t]{0.5\textwidth}
        \begin{center}
        \begin{tabular}{lcrl}
            Method & Bits/dim & \multicolumn{2}{c}{Images/sec}\\
            \hline
            Softmax & 0.621 & 0.45 & (1$\times$) \\
            Stateful-softmax & 0.621 & 7.56 & (16.8$\times$) \\
            LSH-1 & 0.745 & 0.68 & (1.5$\times$) \\
            LSH-4 & 0.676 & 0.27 & (0.6$\times$) \\
            \hline
            Linear (ours) & 0.644 & \textbf{142.8} & \textbf{(317$\times$)}
        \end{tabular}
        \end{center}
        \caption{Image generation on MNIST}
        \label{tab:mnist-rnn}
    \end{subtable}
    \begin{subtable}[t]{0.5\textwidth}
        \begin{center}
        \begin{tabular}{lcrl}
            Method & Bits/dim & \multicolumn{2}{c}{Images/sec} \\
            \hline
            Softmax & 3.47 & 0.004 & (1$\times$) \\
            Stateful-softmax & 3.47 & 0.32 & (80$\times$) \\
            LSH-1 & 3.39 & 0.015  & (3.75$\times$) \\
            LSH-4 & 3.51 & 0.005  & (1.25$\times$) \\
            \hline
            Linear (ours) & 3.40 & \textbf{17.85} & \textbf{(4,462$\times$)}
        \end{tabular}
        \end{center}
        \caption{Image generation on CIFAR-10}
        \label{tab:cifar-rnn}
    \end{subtable}
    \caption{Comparison of autoregressive image generation throughput of MNIST
             and CIFAR-10 images. The experiment can be found in \S~4.2 in the
             main paper. For stateful-softmax we save the keys and values and
             reuse them for predicting the next element. A detailed description
             of this extra baseline can be found in
             \S~\ref{sec:imgen-discussion-1}.}
    \label{tab:rnn}
\end{table}

Table \ref{tab:rnn} summarizes the results. We observe that stateful-softmax is
significantly faster than vanilla transformers. However, its complexity is
still quadratic with respect to the sequence length and our forumlation is more
than 50$\times$ faster for CIFAR-10. Moreover, we would like to point out that
implementing a similar stateful attention for Reformer is not a trivial task as
the sorting and chunking operations need to be performed each time a new input
is provided.

\subsection{Equalizing the batch size} \label{sec:imgen-discussion-2}

In the previous sections we evaluate the throughput of all transformer variants
for the task of autoregressive image generation. However, another important
factor to consider is latency, namely the total time required to produce a
single image. To this end, we use a batch size of 1 and measure the time
required by all methods to generate a single image. In addition to running the
inference on the GPU, we also evaluate the time required on CPU. The results
are reported in table \ref{tab:rnn-single}.

\vspace{1em}
\begin{table}[h]
    \begin{subtable}[t]{0.5\textwidth}
        \begin{center}
        \resizebox{0.95\textwidth}{!}{
        \begin{tabular}{lrlrl}
            Method & \multicolumn{2}{c}{Seconds (CPU)} &
                \multicolumn{2}{c}{Seconds (GPU)}\\
            \hline
            Softmax & 72.6 & (13.2$\times$) & 10.2 & (1.4$\times$) \\
            Stateful-softmax & 7.4 & (1.3$\times$) & 10.4 & (1.42$\times$) \\
            LSH-1 & 46.0 & (8.3$\times$) & 19.2 & (2.6$\times$) \\
            LSH-4 & 112.0 & (20$\times$) & 55.8 & (7.6$\times$) \\
            \hline
            Linear (ours) & \textbf{5.5} & (1$\times$) & \textbf{7.3} &
                (1$\times$) \\
        \end{tabular}}
        \end{center}
        \caption{Image generation on MNIST}
        \label{tab:mnist-rnn}
    \end{subtable}    \begin{subtable}[t]{0.5\textwidth}
        \begin{center}
        \resizebox{0.99\textwidth}{!}{
        \begin{tabular}{lrlrl}
            Method & \multicolumn{2}{c}{Seconds (CPU)} &
                \multicolumn{2}{c}{Seconds (GPU)}\\
            \hline
            Softmax & 8651.4 & (191.8$\times$) & 300.1 & (4.9$\times$) \\
            Stateful-softmax & 71.9 & (1.6$\times$) & 70.4 & (1.14$\times$) \\
            LSH-1 & 2318.9 & (51.4$\times$) & 221.6 & (3.6$\times$) \\
            LSH-4 & 5263.7 & (116.7$\times$) & 683.9 & (11.1$\times$) \\
            \hline
            Linear (ours) & \textbf{45.1} & (1$\times$)  & \textbf{61.3} &
                (1$\times$) \\
        \end{tabular}}
        \end{center}
        \caption{Image generation on CIFAR-10}
        \label{tab:cifar-rnn}
    \end{subtable}
    \caption{Comparison of the time required to generate a single image with
             autoregressive transformers on MNIST and CIFAR-10. We run all
             methods with a batch size of 1 both on CPU and GPU and report the
             total time in seconds. For all numbers in the table, lower is
             better.}
    \label{tab:rnn-single}
\end{table}

We observe that all methods underutilize the GPU and achieve significantly
smaller image generation throughput than the one shown in table \ref{tab:rnn}.
The proposed linear transformer is faster than all the methods and in
particular it is almost 6.6$\times$ faster than softmax transformers for
generating an image on CIFAR-10. Note that our linear autoregressive
transformer is the only method that is faster on the CPU than on the GPU in
every case. This is due to the fact that computing the attention as an RNN has
such a low cost that the main computational bottleneck becomes the inevitable
outer loop over the sequence.

\section{Qualitative Results on Image Generation}

In this section we provide qualitative results for our image generation
experiments. Since the perplexity of all models is approximately the same, as
expected, the qualitative differences are not significant. A rather interesting
observation however is that the Reformer models provide significantly fewer
variations in their unconditional samples. Moreover, we observe that image
completion is a significantly easier task than unconditional generation as all
models perform significantly better.

\begin{figure*}[h]
    \centering
    \begin{subfigure}[t]{0.4\columnwidth}
        \includegraphics[width=\linewidth]{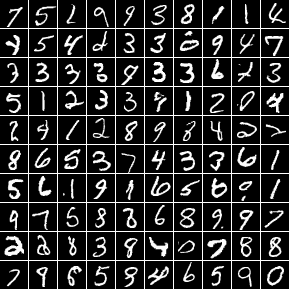}
        \caption{Softmax}
    \end{subfigure}
    \quad
    \begin{subfigure}[t]{0.4\columnwidth}
        \includegraphics[width=\linewidth]{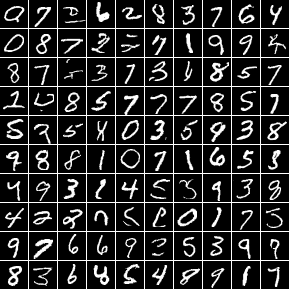}
        \caption{Linear (ours)}
    \end{subfigure}\\[1em]
    \begin{subfigure}[t]{0.4\columnwidth}
        \includegraphics[width=\linewidth]{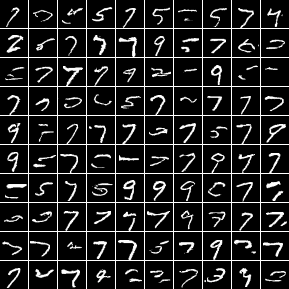}
        \caption{LSH-1}
    \end{subfigure}
    \quad
    \begin{subfigure}[t]{0.4\columnwidth}
        \includegraphics[width=\linewidth]{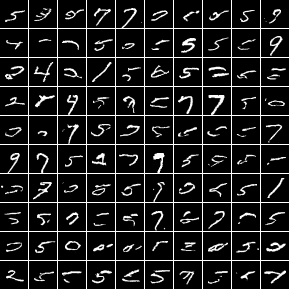}
        \caption{LSH-4}
    \end{subfigure}
    \caption{Unconditional samples from the transformer models trained with
             MNIST. See \S~{4.2.1} in the main paper.}
    \label{fig:cifar_unconditional}
\end{figure*}

\begin{figure*}[h]
    \centering
    \begin{subfigure}[t]{0.15\columnwidth}
        \centering
        \includegraphics[width=0.33\linewidth]{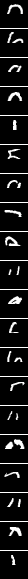}
        \caption{Occluded}
    \end{subfigure}
    \begin{subfigure}[t]{0.15\columnwidth}
        \centering
        \includegraphics[width=0.33\linewidth]{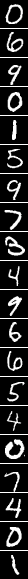}
        \caption{Softmax}
    \end{subfigure}
    \begin{subfigure}[t]{0.15\columnwidth}
        \centering
        \includegraphics[width=0.33\linewidth]{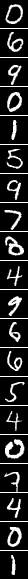}
        \caption{Linear (ours)}
    \end{subfigure}
    \begin{subfigure}[t]{0.15\columnwidth}
        \centering
        \includegraphics[width=0.33\linewidth]{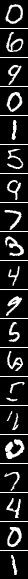}
        \caption{LSH-1}
    \end{subfigure}
    \begin{subfigure}[t]{0.15\columnwidth}
        \centering
        \includegraphics[width=0.33\linewidth]{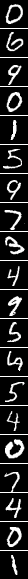}
        \caption{LSH-4}
    \end{subfigure}
    \begin{subfigure}[t]{0.15\columnwidth}
        \centering
        \includegraphics[width=0.33\linewidth]{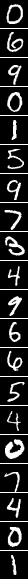}
        \caption{Original}
    \end{subfigure}
    \caption{MNIST digit completion from all trained models.
             See \S~{4.2.1} in the main paper.}
    \label{fig:cifar_unconditional}
\end{figure*}

\begin{figure*}[h]
    \centering
    \begin{subfigure}[t]{0.4\columnwidth}
        \includegraphics[width=\linewidth]{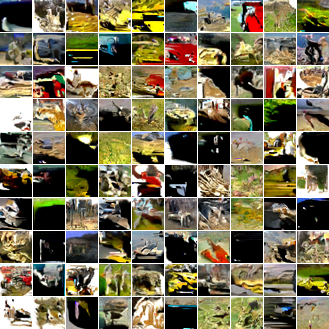}
        \caption{Softmax}
    \end{subfigure}
    \quad
    \begin{subfigure}[t]{0.4\columnwidth}
        \includegraphics[width=\linewidth]{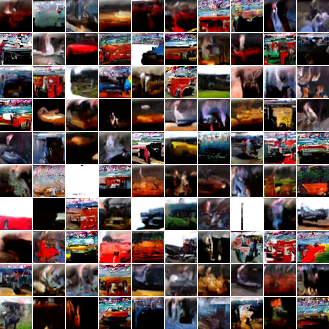}
        \caption{Linear (ours)}
    \end{subfigure}\\[1em]
    \begin{subfigure}[t]{0.4\columnwidth}
        \includegraphics[width=\linewidth]{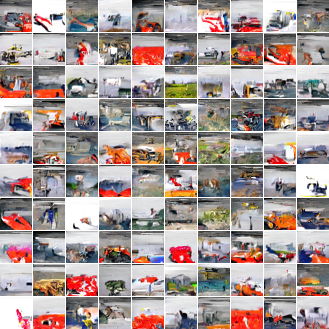}
        \caption{LSH-1}
    \end{subfigure}
    \quad
    \begin{subfigure}[t]{0.4\columnwidth}
        \includegraphics[width=\linewidth]{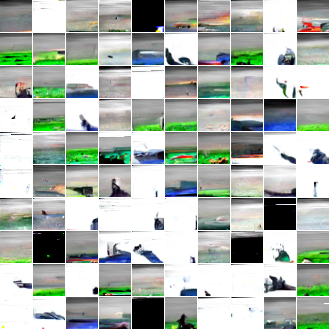}
        \caption{LSH-4}
    \end{subfigure}
    \caption{Unconditional samples from the transformer models trained with
             CIFAR-10. See \S~{4.2.2} in the main paper.}
    \label{fig:cifar_unconditional}
\end{figure*}

\begin{figure*}[h]
    \centering
    \begin{subfigure}[t]{0.15\columnwidth}
        \centering
        \includegraphics[width=0.33\linewidth]{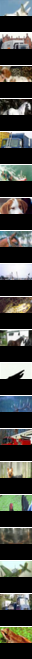}
        \caption{Occluded}
    \end{subfigure}
    \begin{subfigure}[t]{0.15\columnwidth}
        \centering
        \includegraphics[width=0.33\linewidth]{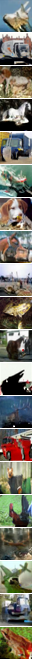}
        \caption{Softmax}
    \end{subfigure}
    \begin{subfigure}[t]{0.15\columnwidth}
        \centering
        \includegraphics[width=0.33\linewidth]{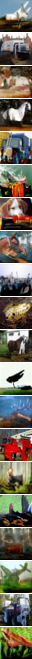}
        \caption{Linear (ours)}
    \end{subfigure}
    \begin{subfigure}[t]{0.15\columnwidth}
        \centering
        \includegraphics[width=0.33\linewidth]{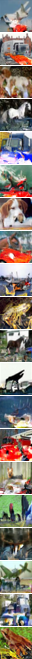}
        \caption{LSH-1}
    \end{subfigure}
    \begin{subfigure}[t]{0.15\columnwidth}
        \centering
        \includegraphics[width=0.33\linewidth]{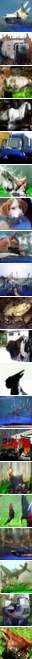}
        \caption{LSH-4}
    \end{subfigure}
    \begin{subfigure}[t]{0.15\columnwidth}
        \centering
        \includegraphics[width=0.33\linewidth]{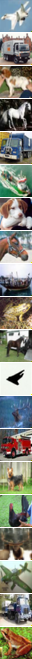}
        \caption{Original}
    \end{subfigure}
    \caption{CIFAR-10 image completions from all trained transformer models.
             See \S~{4.2.2} in the main paper.}
    \label{fig:cifar_unconditional}
\end{figure*}

\checknbdrafts
\end{document}